\documentclass[11pt,a4paper]{article}
\usepackage[hyperref]{acl2019}
\usepackage{times}
\usepackage{latexsym}
\usepackage{array, multirow}
\usepackage{graphicx}
\usepackage{amsfonts}
\usepackage{amsmath}
\usepackage{mathtools}
\usepackage{rotating} 
\usepackage{rotfloat}

\usepackage{url}

\aclfinalcopy 
\def\aclpaperid{2639} 

\title{Multimodal Transformer Networks for End-to-End \\Video-Grounded Dialogue Systems}

\author{Hung Le$^{1,2}$, Doyen Sahoo$^1$, Nancy F. Chen$^2$, Steven C.H. Hoi$^{1,3}$ \\
  $^1$Singapore Management University \\
  $^2$Institute of Inforcomm Research (I2R), Singapore \\
  $^3$Salesforce Research Asia \\
  \texttt{\{hungle.2018,doyens\}@smu.edu.sg}\\
  \texttt{nfychen@i2r.a-star.edu.sg,chhoi@smu.edu.sg}
 }

\date{}

\begin{document}
\maketitle

\begin{abstract}
Developing Video-Grounded Dialogue Systems (VGDS), where a dialogue is conducted based on visual and audio aspects of a given video, is significantly more challenging than traditional image or text-grounded dialogue systems because 
(1) feature space of videos span across multiple picture frames, making it difficult to obtain semantic information; and (2) a dialogue agent must perceive and process information from different modalities (audio, video, caption, etc.) to obtain a comprehensive understanding.  Most existing work is based on RNNs and sequence-to-sequence architectures, which are not very effective for capturing complex long-term dependencies (like in videos).  To overcome this, we propose Multimodal Transformer Networks (MTN) to encode videos and incorporate information from different modalities. We also propose query-aware attention through an auto-encoder to extract query-aware features from non-text modalities. We develop a training procedure to simulate token-level decoding to improve the quality of generated responses during inference. We get state of the art performance on Dialogue System Technology Challenge 7 (DSTC7). Our model also generalizes to another multimodal visual-grounded dialogue task, and obtains promising performance.  We implemented our models using PyTorch and the code is released at \url{https://github.com/henryhungle/MTN}.

\end{abstract}

\section{Introduction}

\begin{figure}[h]
    \centering
	\includegraphics[width=\columnwidth]{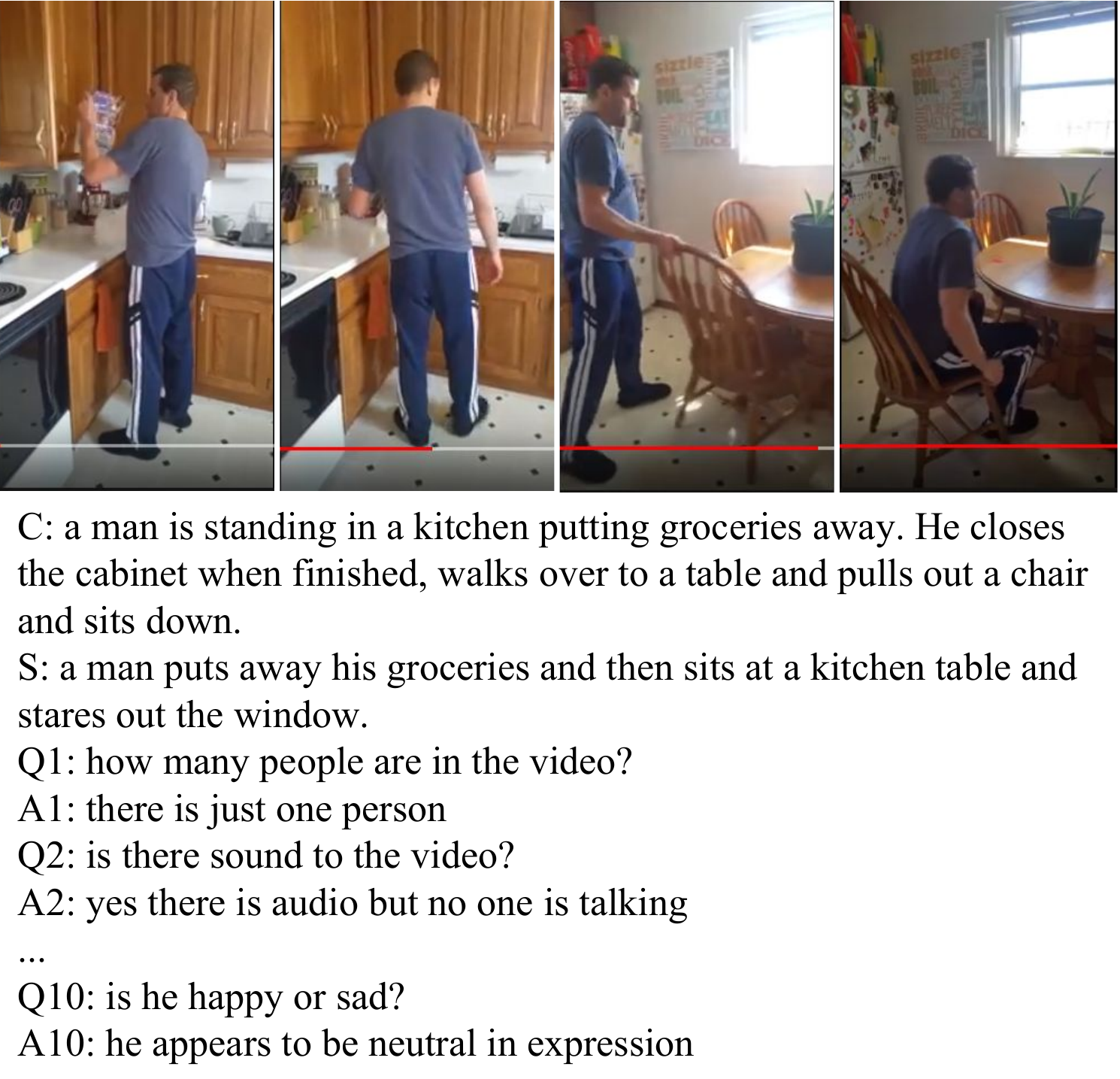}
	\caption{A sample dialogue from the DSTC7 Video Scene-aware Dialogue training set with 4 example video scenes. \textit{C}: Video Caption, \textit{S}: Video Summary, \textit{Qi}: $i^{th}$-turn question, \textit{Ai}: $i^{th}$-turn answer}
	\label{fig:data}
	\vspace{-0.5cm}
\end{figure}

A  video-grounded dialogue system (VGDS) generates appropriate conversational response to queries of humans, by not only keeping track of the relevant dialogue context, but also understanding the relevance of the query in the context of a given video (knowledge grounded in a video) \cite{hori2018end}. An example dialogue exchange can be seen in Figure \ref{fig:data}. Developing such systems has recently received interest from the research community (e.g. DSTC7 challenge \cite{DSTC7}). This task is much more challenging than traditional text-grounded or image-grounded dialogue systems because:
(1) feature space of videos is larger and more complex than text-based or image-based features because of diverse information, such as background noise, human speech, flow of actions, etc.  across multiple video frames; and (2) a conversational agent must have the ability to perceive and comprehend information from different modalities (text from dialogue history and human queries, visual and audio features from the video) and semantically shape a meaningful response to humans.  

Most existing approaches for multi-modal dialogue systems are based on RNNs as the sequence processing unit and sequence-to-sequence network as the overall architecture to model the sequential information in text \cite{das2017visual, das2017learning, hori2018end,kottur2018visual}. Some efforts adopted query-aware attention to allow the models to focus on specific parts of the features most relevant to the dialogue context \cite{hori2018end, kottur2018visual}.
Despite promising results, these methods are not very effective or efficient for processing video-frames, due to the complexity of long term sequential information from multiple modalities.
We propose Multimodal Transformer Networks (MTN) which model the complex sequential information from video frames, and also incorporate information from different modalities. 
MTNs allow for complex reasoning over multimodal data such as in videos, by jointly attending to information in different representation subspaces, and making it easier (than RNNs) to fuse information from different modalities. 
Inspired by the success of Transformers \cite{vaswani17attention}) for text, we propose novel neural architectures for VGDS:
(1) We propose to capture complex sequential information from video frames using multi-head attention layers. Multi-head attention is applied across several modalities (visual, audio, captions) repeatedly. This works like a memory network to allow the models to comprehensively reason over the video to answer human queries;
(2) We propose an auto-encoder component, designed as query-aware attention layer, to further improve the reasoning capability of the models on the non-text features of the input videos; and
(3) We employ a training approach to improve the generated responses by simulating token-level decoding during training. 

We evaluated MTN on a video-grounded dialogue dataset (released through DSTC7 \cite{DSTC7}). In each dialogue, video features such as audio, visual, and video caption, are available, which have to be processed and understood to hold a conversation. We conduct comprehensive experiments to validate our approach, including automatic evaluations, ablations, and qualitative analysis of our results. We also validate our approach on the visual-grounded dialogue task \cite{das2017visual}, and show that MTN can generalize to other multimodal dialog systems. 

\section{Related Work}
The majority of work in dialogues is formulated as either open-domain dialogues \cite{shang2015,vinyals15neural,YaoZP15, li2016diversity,li2016personas,serban17latent,serban16hierarchical} or task-oriented dialogues \cite{henderson2014wordbaseddst, Bordes2016LearningEG,Fatemi2016,liu2017end,lei2018sequicity,madotto2018mem2seq}. 
Some recent efforts develop conversational agents that ground their responses on external knowledge, e.g. online encyclopedias \cite{dinan2018wizard}, social networks, or user recommendation sites \cite{ghazvininejad17knowledge}. The agent generates a response that can relate to the current dialogue context as well as exploit the information source. Recent dialogue systems use Transformer principles \cite{vaswani17attention} for incorporating attention and focus on different dialogue settings, e.g. text-only or response selection settings \cite{zhu2018sdnet,mazare2018training,dinan2018wizard},
These approaches consider the knowledge to be grounded in text, whereas in VGDS, the knowledge is grounded in videos (with multimodal sources of information). 

There are a few efforts in NLP domain, where multimodal information needs to be incorporated for the task.
Popular research areas include image captioning \cite{vinyals2015show,xu2015show}, video captioning \cite{hori2017attention,li2018jointly} and visual question-answering (QA) \cite{antol2015vqa,goyal2017making}. 
Image captioning and video captioning tasks require to output a description sentence about the content of an image or video respectively. This requires the models to be able to process certain visual features (and audio features in video captioning) and generate a reasonable description sentence. Visual QA involves generating a correct response to answer a factual question about a given image. The recently proposed movie QA \cite{MovieQA} task is similar to visual QA but the answers are grounded in movie videos. However, all of these methods are restricted to answering specific queries, and do not maintain a dialogue context, unlike what we aim to achieve in VGDS. We focus on generating dialogue responses rather than selecting from a set of candidates. This requires the dialogue agents to model the semantics of the visual and/or audio contents to output appropriate responses.

Another related task is visual dialogues \cite{das2017visual,das2017learning,kottur2018visual}. This is similar to visual QA but the conversational agent needs to track the dialogue context to generate a response. However, the knowledge is grounded in images. In contrast, we focus on knowledge grounded in videos, which is more complex, considering the large feature space spanning across multiple video frames and modalities that need to be understood. 

\section{Multimodal Transformer Networks}

Given an input video $V$, its caption $C$, a dialogue context of $(t-1)$ turns, each including a pair of (question, answer) $(Q_1, A_1),...,(Q_{t-1},A_{t-1})$, and a factual query $Q_t$ on the video content, the goal of a VGDS is to generate an appropriate dialogue response $A_t$. We follow the attention-based principle of Transformer network \cite{vaswani17attention}  and propose a novel architecture: \textit{Multimodal Transformer Networks} to elegantly fuse feature representations from different modalities. MTN enables complex reasoning over long video sequences by attending to important feature representations in different modalities. 

\begin{figure*}[htbp]
	\centering
	\includegraphics[width=\textwidth]{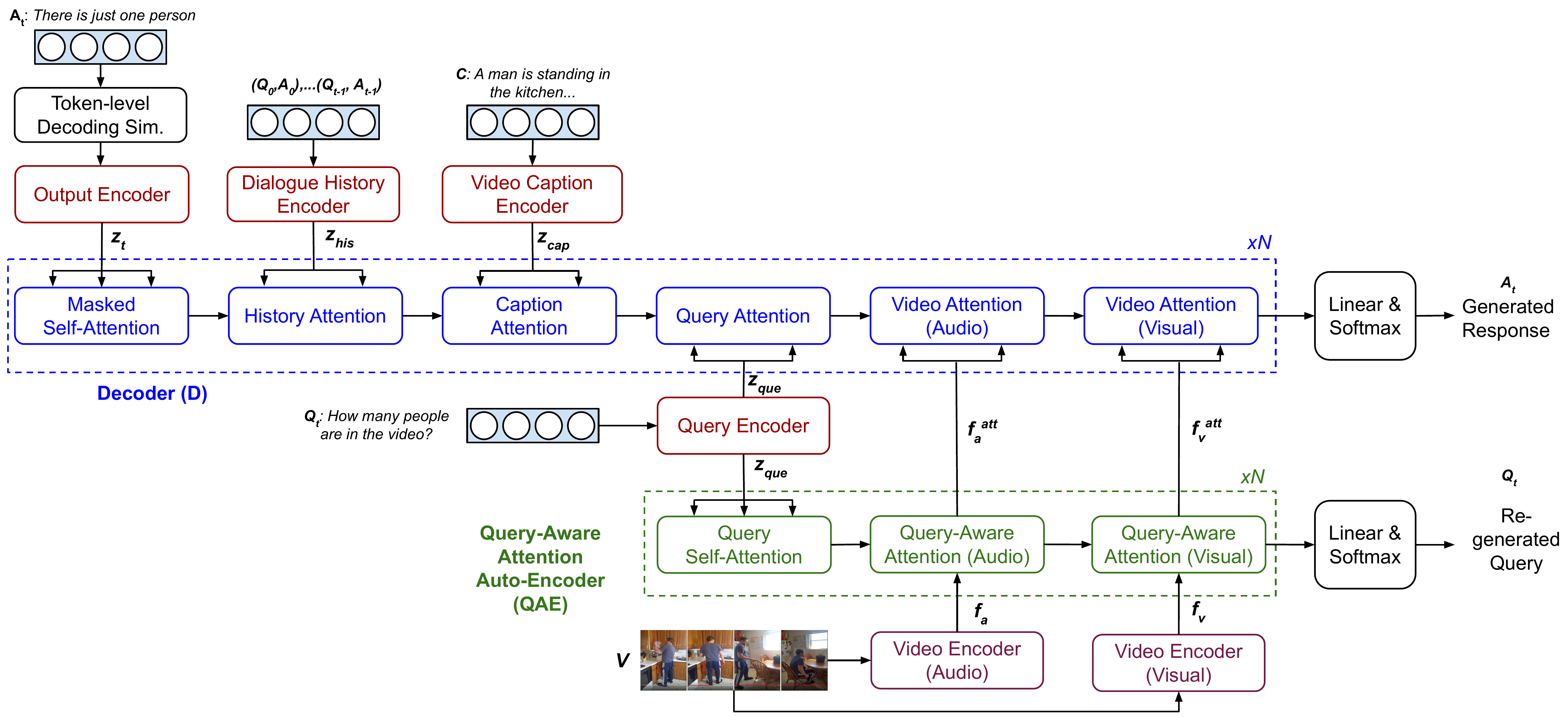}
	\caption{Our MTN architecture includes 3 major components: (i) encoder layers encode text sequences and video features; (ii) decoder layers (D) project target sequence and attend on multiple inputs; and (iii) Query-Aware Auto-Encoder layers (QAE) attend on non-text modalities from query features. For simplicity, Feed Forward, Residual Connection and Layer Normalization layers are not presented. Best viewed in color.}
	\label{fig:model}
\end{figure*}

MTN comprises 3 major components: encoder, decoder, and auto-encoder layers. (i) \textit{Encoder layers} encode text sequences and input video into continuous representations. Positional encoding is used to inject the sequential characteristics of input text and video features at token and video-frame level respectively;  (ii) \textit{Decoder layers} project the target sequences and perform reasoning over multiple encoded features through a multi-head attention mechanism. Attention layers coupled with feed-forward and residual connections process the projected target sequence over $N$ attention steps before passing to a generative component to generate a response; (iii) \textit{Auto-encoder layers} enhance video features with a query-aware attentions on the visual and audio aspects of the input video. A network of multi-head attentions layers are employed as a query auto-encoder to learn the attention in an unsupervised manner. We combine these modules as a Multimodal Transformer Network (MTN) model and jointly train the model end-to-end. An overview of the MTN architecture is shown in Figure \ref{fig:model}. Next, we will discuss the details of each of these components.

\begin{figure*}[!htb]
	\centering
	\includegraphics[width=\textwidth]{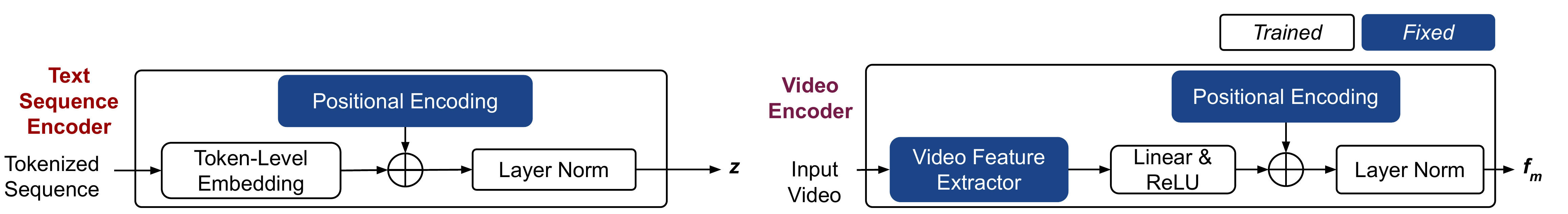}
	\caption{2 types of encoders are used: text-sequence encoders (left) and video encoders (right). Text-sequence encoders are used on text input, i.e. dialogue history, video caption, query, and output sequence. Video encoders are used on visual and audio features of input video.}
	\label{fig:encoders}
\end{figure*}

\subsection{Encoder Layers}
\textbf{Text Sequence Encoders}. The encoder layers map each sequence of tokens $(x_1,...,x_n)$ to a sequence of continuous representation $z=(z_1,...,z_n) \in \mathbb{R}^d$. An overview of text sequence encoder can be seen in Figure \ref{fig:encoders}. The encoder is composed of a token-level learned embedding, a fixed positional encoding layer, and layer normalization. 
We use the positional encoding to incorporate sequential information of the source sequences.  The token-level positional embedding is added on top of the embedding layer by using element-wise summation. Both learned embedding and positional encoding has the same dimension $d$. We used the sine and cosine functions for the positional encoding as similarly adopted in \cite{vaswani17attention}. Compared to a Transformer encoder, we do not use stack of encoder layers with self-attention to encode source sequences. Instead, we only use layer normalization \cite{ba2016layer} on top of the embedding. We also experimented with using stacked Transformer encoder blocks, consisting of self-attention and feed-forward layers, and compare with our approach (see Table \ref{tab:ablation} Row A and B-1). The target sequence $A_t=(y_1,...,y_m)$ is offset by one position to ensure that the prediction in the decoding step $i$ is auto-regressive only on the previously positions $1,...,(i-1)$. Here we share the embedding weights of encoders for source sequences i.e. query, video caption, and dialogue history. 

\textbf{Video Encoders}. For a given video $V$, its features are extracted with a sliding window of \textit{n}-video-frame length. This results in modality feature vector $f_m \in \mathbb{R}^{numSeqs \times d_m}$ for a modality $m$. Each $f_m$ represents the features for a sequence of $n$ video frames. Here we consider both visual and audio features $M=(v,a)$. We use pre-trained feature extractors and keep the weights of the extractors fixed during training. For a set of scene sequences $s_1,...,s_v$, the extracted features for modality $m$ is $f_m=(f_1,...,f_v)$. 
We apply a linear network with ReLU activation to transform the feature vectors from $d_m$- to $d$-dimensional space. We then also employ the same positional encoding as before to inject sequential information into $f_m$. Refer to Figure \ref{fig:encoders} for an overview of video encoder.

\subsection{Decoder Layers}
Given the continuous representation $z_s$ for each source sequence $x_s$ and $z_t$ for the offset target sequence, the decoder generates an output sequence $(y_2,...,y_m)$ (The first token is always an $\langle sos \rangle$ token). The decoder is composed of a stack of $N$ identical layers. Each layer has $4 + \|M\|$ sub-layers, each of which performs attention on an individual encoded input: the offset target sequence $z_t$, dialogue history $z_{his}$, video caption $z_{cap}$, user query $z_{que}$, and video non-text features $\{f_a, f_v\}$. Each sub-layer consists of a multi-head attention mechanism and a position-wise feed-forward layer. Each feed-forward network consists of 2 linear transformation with ReLU activation in between. We employed residual connection \cite{he2016deep} and layer normalization \cite{ba2016layer} around each attention block. The multi-head attention on $z_s$ is defined as: 
\begin{align}
    m_s = Concat(h_1,...,h_h)W^O \\
    h_i = Attn(z^{dec}_{out} W^Q_i, z_s W^K_i, z_s W^V_i) \\
    Attn(q,k,v) = softmax(\frac{qk^T}{\sqrt{d_k}})v 
    \label{equa:multi_head}
\end{align}
where $ W^Q_i\in \mathbb{R}^{d \times d_k}, W^K_i\in \mathbb{R}^{d \times d_k}, W^V_i\in \mathbb{R}^{d \times d_k}, W^O_i\in \mathbb{R}^{hd_v \times d}$ (the superscripts of $s$ and $t$ are not presented for each $W$ for simplicity). $z^{dec}_{out}$ is the output of the previous sub-layer. 

The multi-head attention allows the model to attend on text sequence features at different positions of the sequences. 
By using multi-head attention on visual and audio features, the model can attend on frame sequences to project and extract information from different parts of the video. Using multiple attentions for different input components also allows the model attend differently on inputs rather than using the same attention network for all. We also experimented with concatenating the input sequences and only use one attention block in each decoding layer, similarly to a Transformer decoder ( See the appendix Section \ref{app:additional_results}). 

\subsection{Auto-Encoder Layers}
As the multi-head attentions allow dynamic attentions on different input components, the essential interaction between the input query and non-text features of the input video is not fully implemented. While a residual connection is employed and the video attention block is placed at the end of the decoder layer, the attention on video features might not be optimal. We consider adding query-aware attention on video features as a separate component. We design it as a query auto-encoder to allow the model to focus on query-related features of the video in an unsupervised manner. The auto-encoder is composed of a stack of $N$ layers, each of which includes an query self-attention and query-aware attention on video features. Hence, the number of sub-layers is $1 + \|M\|$. For self-attention, the output of the previous sub-layer $z^{ae}_{out}$ (or $z_{que}$ in case of the first auto-encoder stack) is used identically as $q$, $k$ and $v$ in Equation \ref{equa:multi_head}, while for query-aware attention, $z^{ae}_{out}$ is used as $q$ and $f_m$ is used as $k$ and $v$. For an $n^{th}$ auto-encoder layer, each output of the query-aware attention on video features $f_{m,n}^{att}$ is passed to video attention module of the corresponding $n^{th}$ decoder layer. Each video attention head $i$ for a given modality $m$ at decoding layer $n^{th}$ is defined as: 

\begin{align*}
    h_i &= Attn(z^{dec}_{out,n} W^Q_i, f_{m,n}^{att} W^K_i, f_{m,n}^{att} W^V_i) 
\end{align*}

The decoder and auto-encoder create a network similar to the One-to-Many setting in \cite{luong2015multi} as the encoded query features are shared between the two modules. We also consider using the auto-encoder as stacked query-aware encoder layers i.e. use query self-attention and query-based attention on video features and extract the output of final layer at $N^{th}$ block to the decoder. 
Comparison of the performance (See Table \ref{tab:ablation} Row C-5 and D) shows that adopting an auto-encoder architecture is more effective in capturing relevant video features.

\subsection{Generative Network}
Similar to sequence generative models \cite{Sutskever14seq2seq,Manning2017ACS}, we use a Linear transformation layer with softmax function on the decoder output to predict probabilities of the next token. In the auto-encoder, the same architecture is used to re-generate the query sequence. We separate the weight matrix between the source sequence embedding, output embedding, and the pre-softmax linear transformation. 

\textbf{Simulated Token-level Decoding}. Different from training, during test time, decoding is still an auto-regressive process where the decoder generates the sentence token-by-token. We aim to simulate this process during training by performing the following procedures: 
\begin{itemize}
    \item Rather than always using the full target sequence of length $L$, the token-level decoding simulation will do the following:
    \item With a probability $p$, e.g. $p=0.5$ i.e. for 50\% of time, crop the target sequence at a uniform-randomly selected position $i$ where $i=2,...,(L-1)$ and keep the left sequence as the target sequence e.g. \textit{$\langle sos \rangle $ there is just one person $\langle eos \rangle $} $\rightarrow$ \textit{$\langle sos \rangle$ there is just one}
    \item As before, the target sequence is offset by one position as input to the decoder
\end{itemize}

We employ this approach to reduce the mismatch of input to the decoder during training and test time and hence, improve the quality of the generated responses. We only apply this procedure for the target sequences to the decoder but not the query auto-encoder.

\section{Experiments}

\subsection{Data}
We used the dataset from DSTC7 \cite{DSTC7} which consists of multi-modal dialogues grounded on the Charades videos \cite{sigurdsson2016hollywood}. Table \ref{tab:datasets} summarizes the dataset and Figure \ref{fig:data} shows a training example.
We used the audio and visual feature extractors pre-trained on YouTube videos and the Kinetics dataset \cite{kay2017kinetics} (Refer to \cite{hori2018end} for the detail video features). Specifically we used the 2048-dimensional I3D\_flow features from the ``Mixed\_5c" layer of the I3D network \cite{carreira2017quo} for visual features and  128-dimensional Audio Set VGGish \cite{hershey2017cnn} for audio features. 
We concatenated the provided caption and summary for each video from the DSTC7 dataset as the default video caption \textit{Cap+Sum}. Other data pre-processing procedures are described in the appendix Section \ref{app:data_preprocessing_video}. 

\begin{table}[htbp]
	\centering
	\resizebox{1.0\columnwidth}{!} {
	\begin{tabular}{llll}
		\hline
		& \textbf{Train} & \textbf{Validation} & \textbf{Test}\\ \hline
		\# of Dialogs & 7,659                                 & 1,787                                   & 1,710	\\ \hline
		\# of Turns   & 153,180                               & 35,740                                  & 13,490   \\ \hline
		\# of Words   & 1,450,754                             & 339,006                                 & 110,252	\\ \hline
	\end{tabular}
	}
	\caption{DSTC7 Video Scene-aware Dialogue Dataset}
	\label{tab:datasets}
\end{table}


\subsection{Training}
We use the standard objective function log-likelihood of the target sequence $T$ given the dialogue history $H$, user query $Q$, video features $V$, and video caption $C$. The log-likelihood of re-generated query is also added when QAE is used: 
\begin{align*}
    L &= L(T) + L(Q) \\
      &= \sum_{m} \log P ( y_m | y_{m-1},..., y_1, H,Q,V,C) + \\
      &= \sum_{n} \log P ( x_n^q | x_{n-1}^q,..., x_1^q, Q,V)
\end{align*}
We train MTN models in two settings: Base and Large. The Base parameters are $N=6, h=8, d=512, d_k=d_v=d/h=64$, and the Large parameters are $N=10, h=16, d=1024, d_k=d_v=d/h=64$. The probability $p$ for simulating token-level decoding is 0.5. We trained each model up to 17 epochs. We used the Adam optimizer \cite{kingma2014adam}. The learning rate is varied over the course of training with strategy adopted similarly in \cite{vaswani17attention}. We used \textit{warmup\_steps} as 9660. We employed dropout \cite{srivastava2014dropout} of 0.1 at all sub-layers and embeddings. Label Smoothing \cite{szegedy2016rethinking} is also applied during training. For all models, we select the latest checkpoints that achieve the lowest perplexity on the validation set. We used beam search with beam size 5 an a length penalty 1.0. The maximum output length during inference is 30 tokens. All models were implemented using PyTorch \cite{paszke2017automatic} \footnote{The code is released at \url{https://github.com/henryhungle/MTN}}.


\subsection{Video-Grounded Dialogues}
We compared MTN models with the baseline \cite{hori2018end} and other submission entries to the DSTC7 Track 3. The evaluation includes 4 word-overlapping-based objective measures: BLEU (1 to 4) \cite{papineni2002bleu}, CIDEr \cite{vedantam2015cider}, ROUGE-L \cite{lin2004rouge}, and METEOR \cite{banerjee2005meteor}. The results were computed based on one reference ground-truth response per test dialogue in the test set. 
As can be seen in Table \ref{tab:dstc_result}, both Base- and Large-MTN models outperform the baseline \cite{hori2018end} in all metrics.
Our Large model outperforms the best previously reported models in the challenge across all the metrics. Even our Base model with smaller parameters outperforms most of the previous results, except for \textit{entry1}, which we outperform in BLEU1-3 and METEOR measures. While some of the submitted models to the challenge utilized external data or ensemble techniques \cite{alamri2018audio}, we only use the given training data from the DSTC7 dataset similarly as the baseline \cite{hori2018end}. 

\begin{figure}[htbp]
	\centering
	\includegraphics[width=\columnwidth]{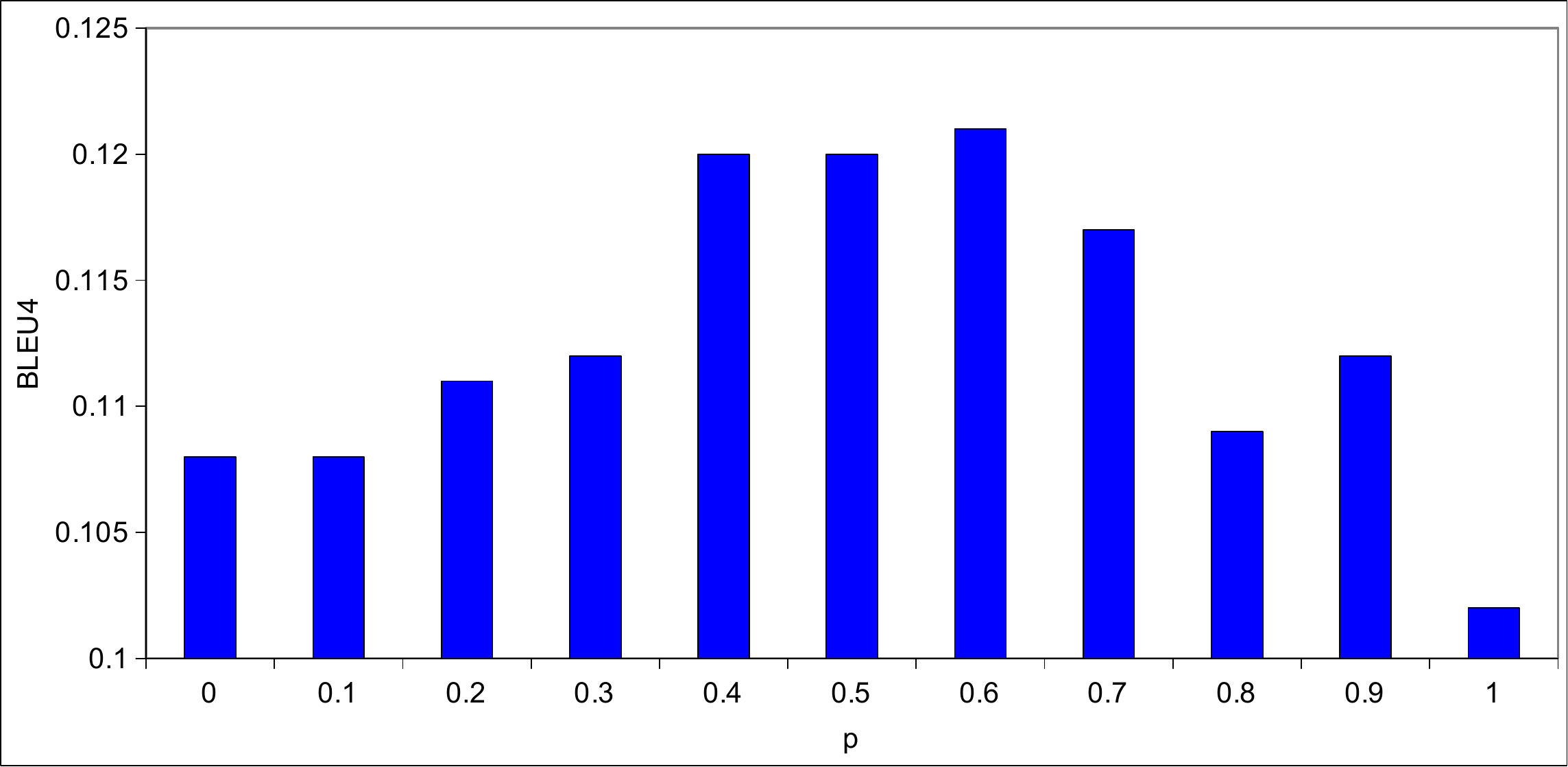}
	\caption{Impact of simulation probability $p$ in BLEU4 measure on the test data. At $p=0.4$ to $0.6$, the improvement in BLEU4 scores is more significant.}
	\label{fig:sim_prob}
\end{figure}

\noindent\textbf{Impact of Token-level Decoding Simulation}. We consider text-only dialogues (no visual or audio features) to study the impact of the token-level decoding simulation component. We also remove the auto-encoder module i.e. \textit{MTN w/o QAE}. We study the differences of performance when the simulation probability $p={0,0.1,...,1}$. 0 is equivalent to always keeping the target sequences as a whole and 1 is cropping all target sequences at random points during training. As shown in Figure \ref{fig:sim_prob}, adding the simulation helps to improve the performance in most cases of $p>0$ and $<1$. 
At $p=1$, the performance is suffered as the decoder receives only fragmented sequences during training. 

\noindent\textbf{Ablation Study}. We tested variants of our models with different combinations of data input in Table \ref{tab:ablation}. With text-only input, compared to our approach (Row B-1), using encoder layers with self-attention blocks (Row A) does not perform well. 
The self-attention encoders also make it hard to optimize the model as noted by \cite{liu2018generating}. When we remove the video caption from the input (hence, no caption attention layers) and use either visual or audio video features, we observe that the proposed auto-encoder with query-aware attention results in better responses. For example, with audio feature, adding the auto-encoder component (Row C-1) increases BLEU4 and CIDEr measures as compared to the case where no auto-encoder is used (Row B-2). When using both caption and video features, the proposed auto-encoder (Row C-5) improves all metrics from the decoder-only model (Row B-4). We also consider using the auto-encoder structure as an encoder (i.e. without the generative component to re-generate query) and decouple from the decoder stacks (i.e. output of the $N^{th}$ encoder layer is used as input to the $1^{st}$ decoder layer) (Row D). The results show that an auto-encoder structure is superior to stacked encoder layers. Our architecture is also better in terms of computation speed as both decoder and auto-encoder are processed in parallel, layer by layer. Results of other model variants are available in the appendix Section \ref{app:additional_results}. 

\subsection{Visual Dialogues}

We also test if MTN could generalize to other multi-modal dialogue settings. We experiment on the visually grounded dialogue task with the VisDial dataset \cite{das2017visual}. The training dataset is much larger than DSTC7 dataset with more than 1.2 million training dialogue turns grounded on images from the COCO dataset \cite{lin2014microsoft}. This task aims to select a response from a set of 100 candidates rather than generating a new complete response. Here we still keep the generative component and maximize the log-likelihood of the ground-truth responses during training. During testing, we use the log-likelihood scores to rank the candidates. We also remove the positional encoding component from the encoder to encode image features as these features do not have sequential characteristics. All other components and parameters remain unchanged. 

We trained MTN with the Base parameters on the Visual Dialogue v1.0 \footnote{\url{https://visualdialog.org/data}} training data and evaluate on the \textit{test-std} v1.0 set. The image features are extracted by a pre-trained object detection model (Refer to the appendix Section \ref{app:data_preprocessing_visual} for data pre-processing).  
We evaluate our model with Normalized Discounted Cumulative Gain (NDCG) score by submitting the predicted ranks of the response candidates to the evaluation server (as the ground-truth for the \textit{test-std} v1.0 split is not published). We keep all the training procedures unchanged from the video-grounded dialogue task. Table \ref{tab:result_vis} shows that our proposed MTN is able to generalize to the visually grounded dialogue setting. It is interesting that our generative model outperforms other retrieval-based approaches in NDCG without any task-specific fine-tuning. There are other submissions with higher NDCG scores from the leaderboard \footnote{\url{https://evalai.cloudcv.org/web/challenges/challenge-page/103/leaderboard/298}} but the approaches of these submissions are not clearly detailed to compare with. 

\begin{table}[htbp]
\centering
\begin{tabular}{ll}
\hline
\textbf{Model} & \textbf{NDCG} \\ \hline
MTN (Base)           & \textbf{55.33}         \\ \hline
CorefNMN \cite{kottur2018visual}       & 54.70         \\ 
MN \cite{das2017visual}            & 47.50         \\ 
HRE \cite{das2017visual}         & 45.46         \\ 
LF \cite{das2017visual}            & 45.31         \\ \hline
\end{tabular}
\caption{Comparison of MTN (Base) to state-of-the-art visual dialogue models on the \textit{test-std} v1.0.  The best measure is highlighted in bold.}
\label{tab:result_vis}
\end{table}

\begin{table*}[htbp]
    \centering
    \resizebox{1.0\textwidth}{!} {
    \begin{tabular}{llllllll}
    \hline
            & \textbf{BLEU1} & \textbf{BLEU2} & \textbf{BLEU3} & \textbf{BLEU4} & \textbf{METEOR} & \textbf{ROUGE-L} & \textbf{CIDEr} \\ \hline
            \multicolumn{8}{l}{\textbf{MTN}}                                          \\ \hline
    MTN (Base)  & \textbf{0.357} & 0.241          & 0.173          & 0.128          & 0.162           & 0.355            & 1.249          \\ 
    MTN (Large) & 0.356          & \textbf{0.242} & \textbf{0.174} & \textbf{0.135} & \textbf{0.165}  & \textbf{0.365}   & \textbf{1.366} \\ \hline
             \multicolumn{8}{l}{\textbf{DSTC7 submissions}}                                                                          \\ \hline
    Entry-top1  & 0.331          & 0.231          & 0.171          & 0.131          & 0.157           & 0.363            & 1.360          \\ 
    Entry-top2  & 0.329          & 0.228          & 0.167          & 0.126          & 0.154           & 0.357            & 1.306          \\ 
    Entry-top3  & 0.327          & 0.225          & 0.164          & 0.123          & 0.155           & 0.350            & 1.269          \\ 
    Entry-top4  & 0.312          & 0.210          & 0.152          & 0.115          & 0.148           & 0.357            & 1.271          \\ 
    Entry-top5  & 0.329          & 0.216          & 0.153          & 0.114          & 0.140           & 0.331            & 1.103          \\ 
   \cite{hori2018end}         & 0.279          & 0.183          & 0.13           & 0.095          & 0.122           & 0.303            & 0.905          \\ \hline
    \end{tabular}
    }
    \caption{Evaluated on the test data, the proposed approach achieves better objective measures than the baselines and the submissions to the challenge. The best result in each metric is highlighted in bold.}
    \label{tab:dstc_result}
\end{table*}

\begin{table*}[htbp]
	\centering
	\resizebox{1.0\textwidth}{!} {
	\begin{tabular}{p{0.1cm}p{1.2cm}llllllll}
\hline
                                         & \textbf{CapFea}       & \textbf{VidFea} & \textbf{BLEU1} & \textbf{BLEU2} & \textbf{BLEU3} & \textbf{BLEU4} & \textbf{METEOR} & \textbf{ROUGE-L} & \textbf{CIDEr} \\ \hline
                                         & \multicolumn{9}{l}{\textbf{MTN w/o QAE + Stacked Self-Attention in Encoder}}                                                                                                  \\ \hline
A                                      & Cap+Sum               & N/A             & 0.327          & 0.216          & 0.154          & 0.114          & 0.147           & 0.332            & 1.106          \\ \hline
                                         & \multicolumn{9}{l}{\textbf{MTN w/o QAE}}   \\ \hline
\multicolumn{1}{c}{B-1} & Cap+Sum               & N/A             & 0.346          & 0.231          & 0.164          & 0.120          & 0.158           & 0.344            & 1.176          \\ 
\multicolumn{1}{c}{B-2}                     & N/A                   & A               & 0.316          & 0.207          & 0.145          & 0.105          & 0.138           & 0.315            & 0.963          \\  
\multicolumn{1}{c}{B-3}                     & N/A                   & V               & 0.328          & 0.222          & 0.158          & 0.118          & 0.147           & 0.331            & 1.102          \\ 
\multicolumn{1}{c}{B-4}                     & Cap+Sum               & A+V             & 0.347          & 0.234          & 0.168          & 0.124          & 0.158           & 0.344            & 1.197          \\ \hline
                                         & \multicolumn{9}{l}{\textbf{MTN}}                                                                                    \\ \hline
\multicolumn{1}{c}{C-1}                     & N/A                   & A               & 0.324          & 0.214          & 0.152          & 0.113          & 0.142           & 0.326            & 1.031          \\ 
\multicolumn{1}{c}{C-2}                                         & N/A                   & V               & 0.328          & 0.223          & 0.155          & 0.119          & 0.147           & 0.330            & 1.115          \\ 
\multicolumn{1}{c}{C-3}                                         & Cap+Sum               & A               & 0.344          & 0.236          & 0.170          & 0.127          & 0.159           & 0.354            & 1.220          \\ 
\multicolumn{1}{c}{C-4}                                         & Cap+Sum               & V               & 0.343          & 0.229          & 0.161          & 0.118          & 0.160           & 0.348            & 1.151          \\  
\multicolumn{1}{c}{C-5}                                         & Cap+Sum               & A+V             & \textbf{0.357}          & \textbf{0.241}          & \textbf{0.173}          & \textbf{0.128}          & \textbf{0.162}           & \textbf{0.355}            & \textbf{1.249}          \\ \hline
                                         & \multicolumn{9}{l}{\textbf{MTN (replacing QAE with QE - Query-Aware Encoder)}}                                                                                         \\ \hline
D                                      & Cap+Sum               & A+V             & 0.334          & 0.227          & 0.164          & 0.123          & 0.153           & 0.344            & 1.200          \\ \hline
\end{tabular}
    }
	\caption{Ablation analysis of MTN evaluated on the test data. The video features being used is either VGGish for audio features (A) or I3D-Flow for visual features (V). All models are trained with the Base parameters. Best result in each metric is highlighted in bold.}
	\label{tab:ablation}
\end{table*}

\section{Qualitative Analysis}

\begin{figure}[htbp]
	\centering
	\includegraphics[width=\columnwidth]{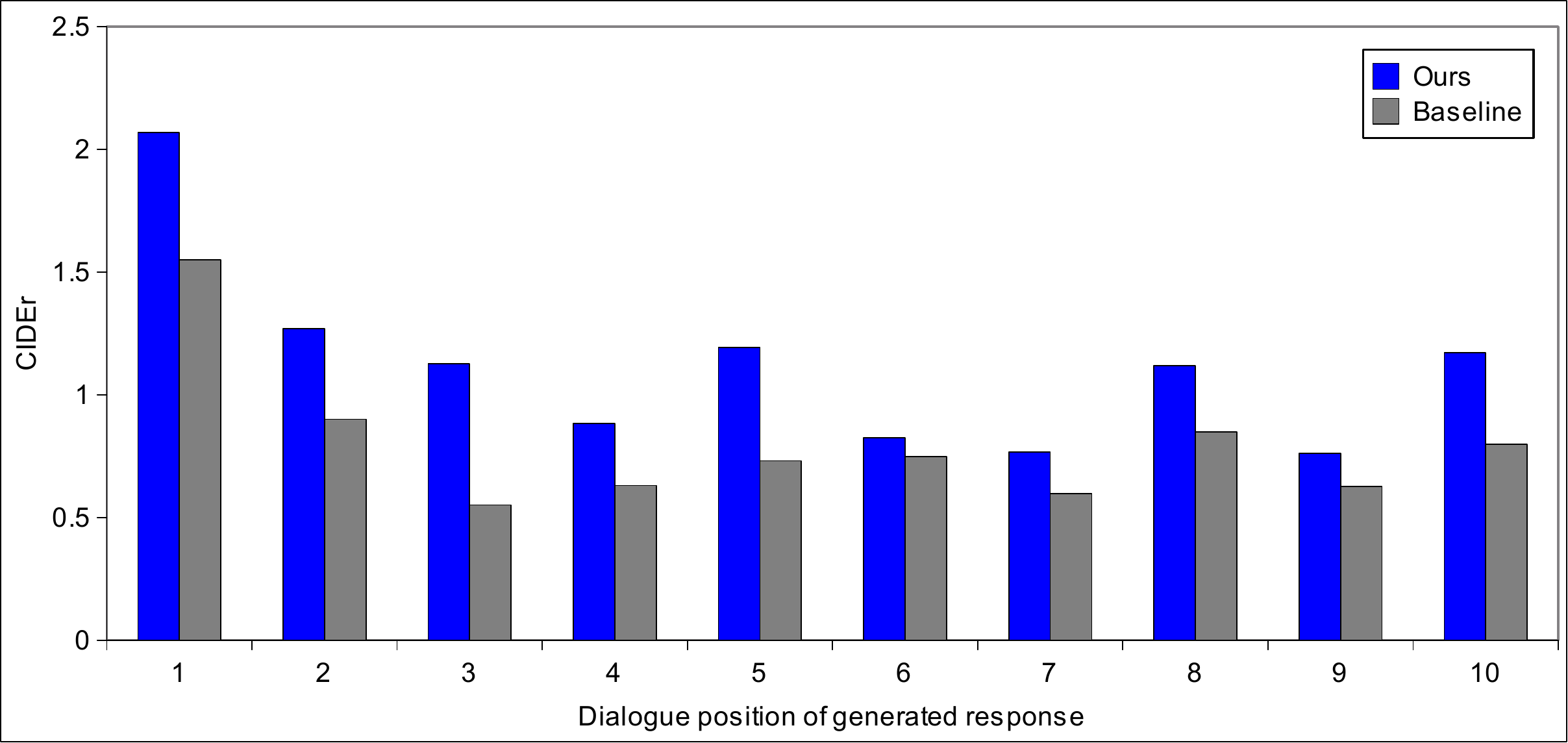}
	\caption{Comparison of CIDEr measures on the test data between MTN (Base) and the baseline \cite{hori2018end} across different turn position of the generated responses. Our model outperforms the baselines at all dialogue turn positions.}
	\label{fig:cider_turns}
\end{figure}

\begin{figure*}[htbp]
	\centering
	\resizebox{1.0\textwidth}{!} {
	\includegraphics{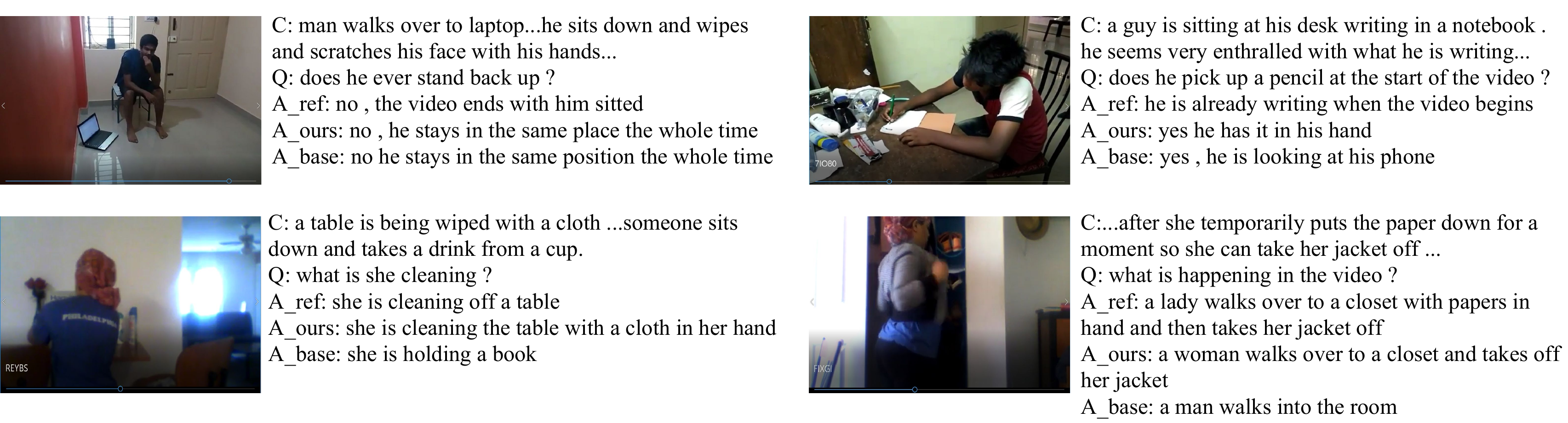}
	}
	\caption{Example test dialogue responses extracted from the ground-truth $A_{ref}$ and generated by MTN (Base) $A_{ours}$ and the baseline \cite{hori2018end} $A_{base}$. For simplicity, the dialogue history is not presented and only parts of the video caption $C$ are shown. Our model provides answers that are more accurate than the baseline, capturing single human action or a series of actions in the videos.}
	\label{fig:samples}
\end{figure*}

Figure \ref{fig:samples} shows some samples of the predicted test dialogue responses of our model as compared to the baseline \cite{hori2018end}. Our generated responses are more accurate than the baseline to answer human queries. Some of our generated responses are more elaborate e.g. ``with a cloth in her hand". Our responses can correctly describe single actions (e.g. ``cleaning the table", ``stays in the same place") or a series of actions (e.g. ``walks over to a closet and takes off her jacket"). This shows that our MTN approach can reason over complex features came from multiple modalities. Figure \ref{fig:cider_turns} summarizes the CIDEr measures of the responses generated by our Base model and the baseline \cite{hori2018end} by their position in dialogue e.g. $1^{st}...10^{th}$ turn. It shows that our responses are better across all dialogue turns, from $1^{st}$ to $10^{th}$. Figure \ref{fig:cider_turns} also shows that MTN perform better at shorter dialogue lengths e.g. 1-turn, 2-turn and 3-turn, in general and the performance could be further improved for longer dialogues. 

\section{Conclusion}
In this paper, we showed that MTN, a multi-head attention-based neural network, can generate good conversational responses in multimodal settings. 
Our MTN models outperform the reported baseline and other submission entries to the DSTC7. We also adapted our approach to a visual dialogue task and achieved excellent performance. 
A possible improvement to our work is adding pre-trained embedding such as BERT \cite{devlin2018bert} or image-grounded word embedding \cite{kiros2018illustrative} to improve the semantic understanding capability of the models. 

\section*{Acknowledgements}
The first author is supported by A*STAR Computing and Information Science scholarship (formerly A*STAR Graduate scholarship). The third author is supported by the Agency for Science, Technology and Research (A*STAR) under its AME Programmatic Funding Scheme (Project \#A18A2b0046).

\bibliography{acl2019}
\bibliographystyle{acl_natbib}

\appendix 

\section{Data Pre-processing}
\subsection{Video-Grounded Dialogues}
\label{app:data_preprocessing_video}
We split all sequences into (case-insensitive) tokens and selected those in the training data with the frequency more than 1 to build the vocabulary for embeddings. This results in 6175 unique tokens, including the $\langle eos \rangle, \langle sos \rangle, \langle pad \rangle $, and $ \langle unk \rangle$ tokens. 
Sentences are batched together by approximate sequence lengths, in order of dialogue history length, video caption length, question length, and target sequence length. We use batch size of 32 during training. 

\subsection{Visual-Grounded Dialogues}
\label{app:data_preprocessing_visual}
The \textit{test-std} v1.0 set include about 4000 dialogues grounded on COCO-like images collected from Flickr. We only selected tokens that have frequency at least 3 in the training data to build the vocabulary. This results in 13832 unique tokens. We use bottom-up attention features \cite{anderson2018bottom} extracted from Faster R-CNN \cite{ren2015faster} which is pre-trained on the Visual Genome data \cite{krishna2017visual}. This results in 36 2048-dimensional feature vectors per image. 

\section{Additional Experiment Results}
\label{app:additional_results}
We experimented our models with text-only input e.g. no video audio or visual features and hence, no auto-encoder layers involved (\textit{MTN w/o QAE}). We tested cases where the maximum dialogue history length $L^{max}_{his}$ is limited to 1, 2, or 3 turns only. For each case, we also tried to concatenate all the source sequences, including dialogue history, video caption, and query, into a single sequence and use only one multi-head attention block on this concatenated sequence in each decoding layer (Similar to a Transformer decoder). Table \ref{tab:his_len} summarizes the results. The results show that concatenating the sequences into one affects the quality of the generated responses significantly. When the input sequences are separated and attended differently by different attention modules, the results improve. This could be explained as different sequences contain different signals to generate responses e.g. dialogue history contains information of references or ellipses in the user queries, user queries include direct signals for feature attention in input videos. 
Another observation is using all possible dialogue turns in the dialogue history i.e. $L^{max}_{his}=10$ achieves the best results. We did not conduct experiments of concatenating source sequences with $L^{max}_{his}=10$ due to memory issues with large input sequences.

\begin{table}[htbp]
	\centering
	\resizebox{1.0\columnwidth}{!} {
	\begin{tabular}{lllll}
		\hline
		\textbf{\begin{tabular}[c]{@{}l@{}}Max.\\ HisLen\end{tabular}} & \textbf{\begin{tabular}[c]{@{}l@{}}Concat. \\ Source \\ Sequence?\end{tabular}} & \textbf{BLEU4} & \textbf{ROUGE-L} & \textbf{CIDEr} \\ \hline
		10& No & \textbf{0.120}   & \textbf{0.344}   & \textbf{1.176} \\ \hline
		3 & No   & 0.116        & 0.343            & 1.141          \\ 
		3 & Yes   & 0.097           & 0.308            & 0.924          \\ \hline
		2 & No     & 0.115      & 0.343            & 1.150          \\ 
		2  & Yes    & 0.090              & 0.304            & 0.900          \\ \hline
		1  & No    & 0.119                    & 0.343            & 1.163          \\ 
		1   & Yes  & 0.095                & 0.301            & 0.894          \\ \hline
	\end{tabular}
	}
	\caption{Evaluation results on the test set for \textit{MTN w/o QAE} models in which maximum history length is range from 1 to 3 or 10 (i.e. all dialogue turns possible). We also experiments when all the source sequences are concatenated into one and the decoder only has one attention block on the concatenated sequence. The auto-encoder components are also removed. Best result in each metric is highlighted in bold.}
	\label{tab:his_len}
\end{table}

\end{document}